\documentclass{article}

\usepackage{PRIMEarxiv}
\usepackage{multirow}
\usepackage{amsmath}
\usepackage{comment}
\usepackage[utf8]{inputenc} % allow utf-8 input
\usepackage[T1]{fontenc}    % use 8-bit T1 fonts
\usepackage{hyperref}       % hyperlinks
\usepackage{url}            % simple URL typesetting
\usepackage{booktabs}       % professional-quality tables
\usepackage{amsfonts}       % blackboard math symbols
\usepackage{nicefrac}       % compact symbols for 1/2, etc.
\usepackage{microtype}      % microtypography
\usepackage{lipsum}
\usepackage{fancyhdr}       % header
\usepackage{graphicx}       % graphics
\graphicspath{{media/}}     % organize your images and other figures under media/ folder

\title{Can ChatGPT Diagnose Alzheimer’s Disease?}

\author{
  Quoc-Toan Nguyen, Linh Le, Xuan-The Tran, Thomas Do, Chin-Teng Lin \\
  GrapheneX-UTS Human-Centric Artificial Intelligence Centre \\ Faculty of Engineering and Information Technology \\ University of Technology Sydney (UTS) \\ Sydney, Australia \\
  \texttt{\{quoctoan.nguyen, linh.le, xuanthe.tran, thomas.do, chin-teng.lin\}@uts.edu.au} \\
}

\begin{document}
\maketitle

\begin{abstract}
Can ChatGPT diagnose Alzheimer’s Disease (AD)? AD is a devastating neurodegenerative condition that affects approximately 1 in 9 individuals aged 65 and older, profoundly impairing memory and cognitive function. This paper utilises 9300 electronic health records (EHRs) with data from Magnetic Resonance Imaging (MRI) and cognitive tests to address an intriguing question: As a general-purpose task solver, can ChatGPT accurately detect AD using EHRs? We present an in-depth evaluation of ChatGPT using a black-box approach with zero-shot and multi-shot methods. This study unlocks ChatGPT's capability to analyse MRI and cognitive test results, as well as its potential as a diagnostic tool for AD. By automating aspects of the diagnostic process, this research opens a transformative approach for the healthcare system, particularly in addressing disparities in resource-limited regions where AD specialists are scarce. Hence, it offers a foundation for a promising method for early detection, supporting individuals with timely interventions, which is paramount for Quality of Life (QoL).
\end{abstract}

% keywords can be removed
\keywords{ChatGPT, Alzheimer's, LLMs, AI in Healthcare, Human-centred Computing}

\section{Introduction}
\label{intro}

Dementia is the seventh most prevalent cause of death globally and is a major contributor to disability and dependence in older adults \cite{who_dementia}. Alzheimer's Disease (AD), the most prevalent form of dementia, is responsible for 60–80\% of cases \cite{2023ADREPORT}, with a high incidence among individuals aged 65 and above \cite{ott1995prevalence,nguyen2024fairad,tran2024eegssmleveragingstatespacemodel,nguyen2024advancing}. AD is characterised by progressive cognitive decline, memory impairment, and neuronal damage, leading to brain atrophy and tissue deterioration \cite{van2023towards}. Although a cure remains unavailable \cite{2023ADREPORT}, early diagnosis plays a critical role in slowing disease progression and enhancing Quality of Life (QoL) through prompt interventions and comprehensive care plans \cite{dubois2016timely, eikelboom2019early}. The progression of AD is typically categorised into three stages \cite{vermunt2019duration}: 1) preclinical, 2) Mild Cognitive Impairment (MCI, also referred to as prodromal AD), and 3) dementia. MCI is characterised by memory deficits but without significant disruptions in daily living activities like dementia \cite{eshkoor2015mild}. 

In recent years, large language models (LLMs) have dramatically advanced in the field of natural language processing (NLP), demonstrating exceptional performance across various NLP tasks \cite{openai2023a,models2023model,bubeck2023sparks,brown2020language,schick2024toolformer,do2024word,cheema2024cd}. Among these, ChatGPT \cite{openai2023a} stands out as a prime example, excelling not only in NLP tasks but also in its ability to follow instructions effectively, generating coherent and informative outputs \cite{jiao2023chatgpt,bang2023multitask,qin2023chatgpt,park2023generative}. Despite their notable capabilities, LLMs may still be hindered by issues of uncertainty, often producing overly confident yet inaccurate responses, a phenomenon is known as `hallucination' \cite{ji2023survey,li2024inference}. Current research predominantly addresses the uncertainty problem in LLMs using a white-box approach. For instance, Kadavath et al. \cite{kadavath2022language} reveal that LLMs are largely aware of their uncertainty by analysing the softmax probabilities. Similarly, Lin et al. \cite{lin2022teaching} highlight that LLMs can be trained to articulate their uncertainty verbally through model fine-tuning. Nevertheless, the white-box approach is not practical. Not all users have the ability or would like to do it. Therefore, evaluation using a black-box approach \cite{yuan2024does,chhikara2024few} without accessing model internal states is relevantly vital to support users who are not experts in artificial intelligence.

Regarding the healthcare sector, ChatGPT and its capabilities have been applied and analysed in many research. For example, there is a positive perception of using it to provide educational materials to patients. This has been proven in research by Pasin \textit{et al.} \cite{tangadulrat2023using}. A study by Jonas \textit{et al.} evaluated ChatGPT-4.0's ability to provide health care advice compared to an expert panel of physicians. It demonstrated superior empathy, usefulness, and correctness in written responses, as rated by patients and specialists \cite{armbruster2024doctor}. A systematic review considered 118 research articles about ChatGPT's applications in patient care, medical research, and publishing. ChatGPT demonstrates potential as a clinical assistant, supporting tasks like patient inquiries, note writing, decision-making, and research \cite{garg2023exploring}. ChatGPT has demonstrated notable potential in various medical applications. For differential diagnosis, Hirosawa \textit{et al.} \cite{hirosawa2023diagnostic} found that ChatGPT-3 achieved a high correct diagnosis rate of 93.3\% in 10 differential-diagnosis lists for common complaints. Rao \textit{et al.} \cite{rao2023assessing} assessed ChatGPT on 36 clinical vignettes, showing an overall diagnostic accuracy of 71.7\%. ChatGPT responses aligned well with the American College of Radiology criteria in cancer screening, achieving an 88.9\% correct rate for select-all-that-apply prompts for breast cancer screening \cite{rao2023evaluating}. Especially an exploratory study using data from four cases has demonstrated ChatGPT's potential in diagnosing AD by accurately assessing cases of varying severity, matching the performance of specialists \cite{el2024chatgpt}.

On top of that, a shortage of geriatricians may hinder the detection of communities as well. A statistic conducted by the American Geriatrics Society expected that the demand will be greater than the supply 1.6 times in 2030 \cite{geriatrics_workforce}. Moreover, using area health resources files, it is estimated that 33–45 specialists per 100,000 are required to provide sufficient care for older adults with MCI and AD. Based on these estimates, 34\%–59\% of the older population may face shortages of dementia specialists \cite{liu2024geographic}. Addressing the shortage of dementia specialists is crucial, particularly in resource-limited areas, and automating the diagnostic approach can play a vital role in delivering faster or more efficient processes not only for specialists but also for individuals. ChatGPT shows promise as a supportive tool in this domain. Hence, this research aims to unlock the potential of ChatGPT using zero-shot and multi-shot prompting methods for AD diagnosis, leveraging 9,300 electronic health records with Magnetic Resonance Imaging (MRI) data and cognitive test scores. This paper's results can open opportunities and ideas to foster this technology for AD detection and advance the healthcare system for dementia care. In short, these are the research questions (RQ) addressed in this paper:
\label{RQ}

\begin{itemize}
    \item \textbf{RQ1:} How effectively does ChatGPT perform in diagnosing AD using MRI data and cognitive test scores?
    \item \textbf{RQ2:} Does the inclusion of MRI data, cognitive test scores, or both enhance the diagnostic accuracy of ChatGPT for AD?
    \item \textbf{RQ3:} Which approach yields better diagnostic performance with ChatGPT: a zero-shot method without prior examples or a multi-shot method with ground truth samples?
\end{itemize}

\section{Related Work}
The number of research leveraging ChatGPT for supporting dementia and AD is likely limited; it has been applied and analysed in just some research. Firstly, a recent pilot study by Aguirre \textit{et al.} \cite{aguirre2024assessing} assessed the potential of ChatGPT-3.5 to support dementia caregivers by providing high-quality responses to real-world questions. Using posts from caregivers on Reddit, researchers evaluated ChatGPT's responses across topics like memory loss, aggression, and driving using a formal rating scale. ChatGPT demonstrated consistently high-quality responses, with 78\% scoring 4 or 5 points out of 5, excelling in synthesizing information and offering recommendations. Next, a study comparing 60 dementia-related queries found Google excelled in currency and reliability, while ChatGPT scored higher in objectivity and relevance. ChatGPT had lower readability (mean grade level 12.17, SD 1.94) than Google (9.86, SD 3.47). Response similarity was high for 13 (21.7\%), medium for 16 (26.7\%), and low for 31 (51.6\%) queries \cite{hristidis2023chatgpt}. ChatGPT was developed to interpret the findings of the output of the introduced TriCOAT model by Diego \textit{et al}. In particular, chatGPT has tremendous potential in AD research, such as early detection \cite{thapaleveraging}. A study evaluated ChatGPT's ability to diagnose AD using four samples as cases with MCI and AD. ChatGPT accurately diagnosed these cases, matching the performance of two AD specialists. The findings highlight ChatGPT’s potential as a tool for AD diagnosis \cite{el2024chatgpt}.

Despite its promising potential, ChatGPT's application in AD detection remains underexplored, particularly with a large-scale dataset. This paper aims to open and disclose ChatGPT's capability to accurately diagnose AD, paving the way for its broader adoption in clinical and research settings.

\section{Material}
Publicly available data from the Alzheimer’s Disease Neuroimaging Initiative (ADNI) \cite{adni_data,jack2008alzheimer,al2023ppad} was utilized for this research due to its large amount of samples. We include data from 1480 individuals, comprising 9300 electronic health records (EHRs) with corresponding MRI volumes and cognitive test scores. Medical professionals labeled these records as NC, MCI, or AD. Each participant may have multiple EHRs due to repeated visits. The dataset includes 3577 records labelled as NC, 4590 as MCI, and 1133 as AD. Table \ref{data_ADNI} provides a detailed breakdown of the data, including cognitive test scores and MRI information used in this research.

\begin{table*}[ht]
\caption{Details of ADNI Dataset with Included Features Used for Experiments of This Research.}
\centering

\begin{tabular}{|c|l|p{7cm}|}
\hline
\textbf{Modality}            & \multicolumn{1}{c|}{\textbf{Abbreviation}}           & \multicolumn{1}{c|}{\textbf{Description}}                                   \\ \hline
\multirow{12}{*}{Cognitive Test}  & CDRSB                   & Clinical Dementia Rating-Sum of Boxes                  \\ \cline{2-3} 
                            & ADAS11                  & Alzheimer’s Disease Assessment Scale 11                \\ \cline{2-3} 
                            & ADAS13                  & Alzheimer’s Disease Assessment Scale 13                \\ \cline{2-3} 
                            & ADASQ4                  & Alzheimer’s Disease Assessment Scale Q4                \\ \cline{2-3} 
                            & MMSE                    & Mini-Mental State Examination                          \\ \cline{2-3} 
                            & RAVLT\_im       & Rey Auditory Verbal Learning Test (Immediate Recall)   \\ \cline{2-3} 
                            & RAVLT\_le        & Rey Auditory Verbal Learning Test (Learning)           \\ \cline{2-3} 
                            & RAVLT\_fo      & Rey Auditory Verbal Learning Test (Forgetting)         \\ \cline{2-3} 
                            & RAVLT\_perc\_fo & Rey Auditory Verbal Learning Test (Percent Forgetting) \\ \cline{2-3} 
                            & LDELTOTAL               & Logical Memory Delayed Recall Total                    \\ \cline{2-3} 
                            & TRABSCOR                & Trail Making Test-B                                    \\ \cline{2-3} 
                            & FAQ                     & Functional Activities Questionnaire                    \\ \hline
\multirow{8}{*}{MRI} & Ventricles              & Ventricles Volume                                 \\ \cline{2-3} 
                            & Hippocampus             & Hippocampus Volume                                \\ \cline{2-3} 
                            & WholeBrain              & Whole Brain Volume                                \\ \cline{2-3} 
                            & Entorhinal              & Entorhinal Cortex Volume                          \\ \cline{2-3} 
                            & Fusiform                & Fusiform Gyrus Volume                                         \\ \cline{2-3} 
                            & MidTemp                 & Middle Temporal Artery Volume                                \\ \cline{2-3} 
                            & ICV                     & Intracranial  Volume                                         \\ \cline{2-3} \hline
\end{tabular}
\label{data_ADNI}
\end{table*}

\section{Method}
\begin{figure*}
    \centering
\includegraphics[width=1\linewidth]{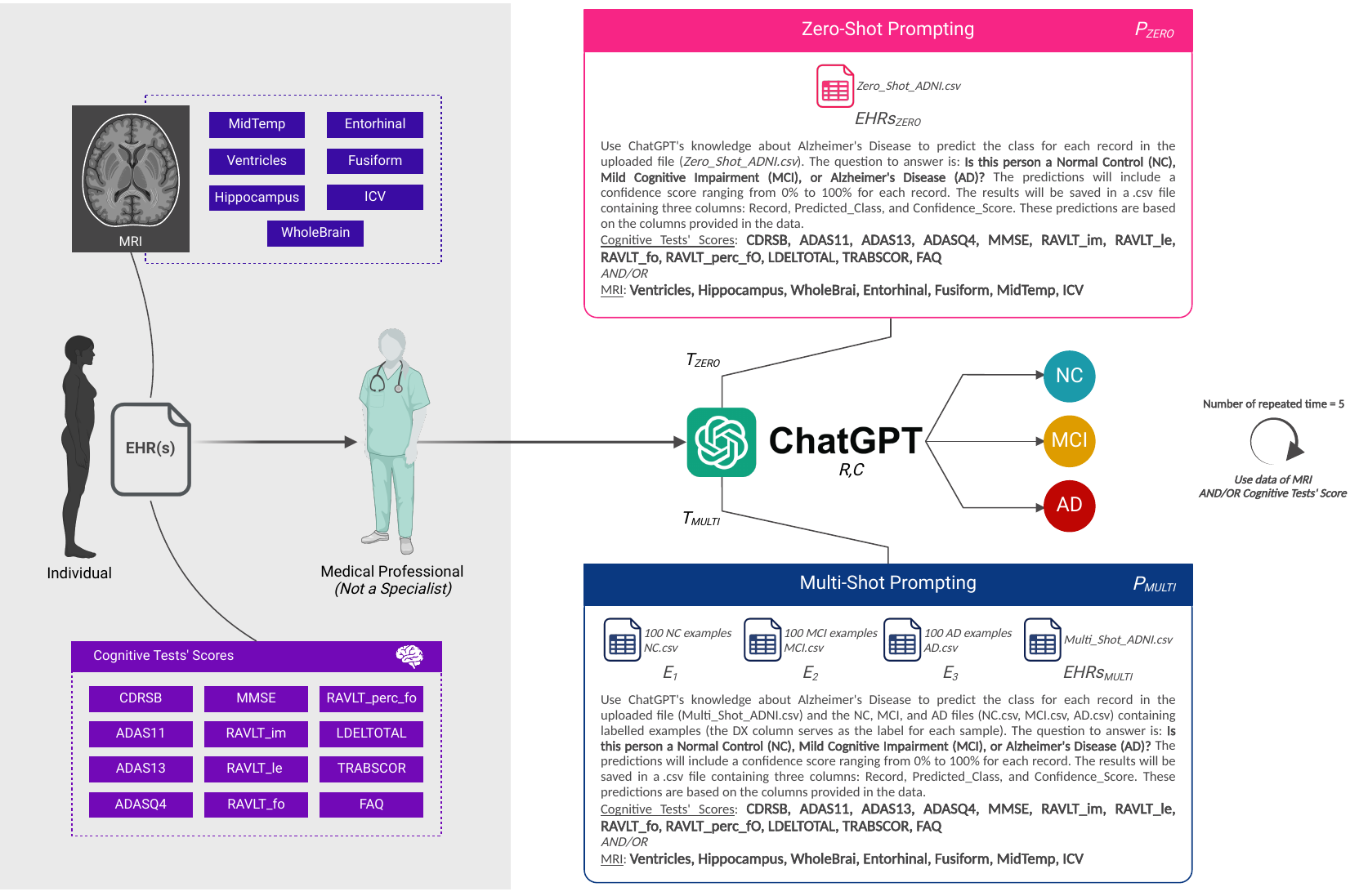}
    \caption{The Workflow of Exploring ChatGPT’s Potential in Diagnosing AD. $P_{ZERO}$ and $P_{MULTI}$ are The Prompts for having the Predictive Results.}
    \label{workflow}
\end{figure*}

In this section, the methods employed in this paper are described in Figure \ref{workflow}. However, before delving into the technical details, it is essential to understand the overarching workflow of this research. The workflow is designed to bridge the gap between real-world applications and the methods studied in this research. On the left, the workflow represents a real-world use case where data from an individual's MRI scans and cognitive tests are either collected or analysed by a medical professional who is not necessarily an AD specialist. This medical staff member can input the available data into ChatGPT, which provides a diagnostic prediction by outputting the individual as NC, MCI, or AD.

On the right, the workflow illustrates how this research is conducted using ChatGPT to diagnose AD. The study explores two distinct prompting approaches—zero-shot prompting and multi-shot prompting—detailed in Sections \ref{zero-shot-prompting} and \ref{multi-shot-prompting}, respectively. Both approaches leverage ChatGPT to predict diagnosis. To ensure the reliability of the outputs, each method is executed five times for MRI or cognitive test scores only, and MRI combined cognitive test scores to evaluate the consistency of ChatGPT's responses.

\subsection{Zero-Shot Prompting}
\label{zero-shot-prompting}
Zero-shot prompting is an approach in NLP where a model is given a task without any task-specific examples \cite{NEURIPS2020_1457c0d6, pmlr-v248-kim24b, o2023accuracy, hu2024zero}. Instead, the task is described directly in the prompt, relying on the model's general knowledge and understanding to generate a response. This method leverages pre-trained models to generalize across tasks without additional fine-tuning. In this paper, a zero-shot learning approach is utilized to develop a prompt using general comprehension of ChatGPT to predict whether the EHR in a CSV file (\( \text{EHRs}_{ZERO} \)) is classified as NC, MCI, or AD based on MRI and/or cognitive test scores. The detail of the proposed prompt is illustrated in Figure \ref{workflow}.

Let \( T_{ZERO} \) represent the task of categorizing NC, MCI, and AD, \( P_{ZERO} \) represent the prompt provided to ChatGPT, \( R \) represent the response generated by the model, and \( C \) represent the confidence score associated with the response. The confidence score quantifies ChatGPT's confidence about the response \( R \).

The zero-shot prompting process can now be described as:

\[
(R, C) = \arg\max_{r, c} \; P(r, c \mid T_{ZERO}, P_{ZERO})
\]

where:
\begin{itemize}
    \item \( r \) is a possible response in the space of all potential outputs,
    \item \( c \) is the associated confidence score for the response \( r \),
    \item \( P(r, c \mid T_{ZERO}, P_{ZERO}) \) is the joint probability of generating a response \( r \) with a confidence score \( c \), given the task \( T_{ZERO} \) and the prompt \( P_{ZERO} \).
\end{itemize}

Breaking this down further, the response \( R \) and its confidence score \( C \) are determined based on the model's ability to evaluate the probability of \( r \) and its confidence \( c \) using pre-trained knowledge \( K \):

\[
P(r, c \mid T_{ZERO}, P_{ZERO}) = g(r, c; T, P_{ZERO}, K)
\]

where:
\begin{itemize}
    \item \( g(r, c; T_{ZERO}, P_{ZERO}, K) \) is a scoring function that the model uses to calculate both the likelihood of the response and the confidence score based on the task, prompt and its pre-trained knowledge,
    \item The confidence score \( C \) is typically derived from the model's internal probability distribution over possible outputs, often normalized to a percentage for interpretability.
\end{itemize}

In the context of classifying EHRs to detect AD, the zero-shot approach generates a predicted class \( R \) (NC, MCI, or AD) and an associated confidence score \( C \), which quantifies the model's confidence about the prediction based on MRI and/or cognitive test scores.

\subsection{Multi-Shot Prompting}
\label{multi-shot-prompting}

Multi-shot prompting leverages multiple example question-and-answer pairs to guide the model. By utilizing these examples, the model may gain a clearer understanding of the intended output \cite{umer2024innovation, zhai2024multi, chun2023explainable, trozze2024large, skilton2024inclusive}. In this paper, examples with ground truth labels of the three classes (NC, MCI, and AD) are provided to improve predictions. Specifically, multi-shot prompting is leveraged to predict EHRs in a CSV file (\( \text{EHRs}_{MULTI} \)) using example files \( E (E_1, E_2, E_3) \) containing ground truth labels. The conducted prompt is presented in Figure \ref{workflow}.

Let \( T_{MULTI} \) represent the task of classifying NC, MCI, and AD, \( P_{MULTI} \) represent the prompt provided to the model, \( E \) represent the set of example question-and-answer pairs, \( R \) represent the response generated by the model, and \( C \) represent the confidence score associated with the response. The confidence score quantifies the model's confidence about the response \( R \).

The multi-shot prompting process can now be described as:

\[
(R, C) = \arg\max_{r, c} \; P(r, c \mid T_{MULTI}, P_{MULTI}, E)
\]

where:
\begin{itemize}
    \item \( r \) is a possible response in the space of all potential outputs,
    \item \( c \) is the associated confidence score for the response \( r \),
    \item \( P(r, c \mid T_{MULTI}, P_{MULTI}, E) \) is the joint probability of generating a response \( r \) with a confidence score \( c \), given the task \( T \), the prompt \( P_{MULTI} \), and the example pairs \( E \).
\end{itemize}

Breaking this down further, the response \( R \) and its confidence score \( C \) are determined based on the model's ability to evaluate the probability of \( r \) and its confidence \( c \) using pre-trained knowledge \( K \) and the examples \( E \):

\[
P(r, c \mid T_{MULTI}, P_{MULTI}, E) = g(r, c; T_{MULTI}, P_{MULTI}, E, K)
\]

where:
\begin{itemize}
    \item \( g(r, c; T_{MULTI}, P_{MULTI}, E, K) \) is a scoring function that the model uses to calculate both the probability of the response and the confidence score based on the task, prompt, examples, and its pre-trained knowledge,
    \item The confidence score \( C \) is typically derived from the model's internal probability distribution over possible outputs, often normalized to a percentage for interpretability.
\end{itemize}

For classifying EHRs, the multi-shot approach utilizes ChatGPT's general knowledge and example question-and-answer pairs E to provide a predicted class \( R \) (NC, MCI, or AD) and an associated confidence score \( C \).

\section{Experiments}
\subsection{Evaluation Metrics}

\label{notation_performance_metrics}

This study employs five essential performance metrics, which are highly relevant for evaluating AI systems in healthcare applications \cite{hicks2022evaluation}: accuracy, recall, precision and F1-score. These metrics are represented as percentages, with values ranging from 0 to 1. A higher value generally indicates better performance across the mentioned metrics.

The calculations of these metrics rely on four foundational components: True Positives (TP), True Negatives (TN), False Positives (FP), and False Negatives (FN). $TP$ refers to the number of positive cases that are correctly identified, while $TN$ represents the number of negative cases correctly classified. $FP$ corresponds to negative cases that are mistakenly classified as positive, and $FN$ accounts for positive cases that are incorrectly labelled as negative. The metrics are computed using the following formulas:

\[ \text{accuracy} = \frac{TP + TN}{TP + FP + TN + FN} \]

\[ \text{recall} = \frac{TP}{TP + FN} \] 

\[ \text{precision} = \frac{TP}{TP + FP} \]

\[ \text{F1-score} = 2 \times \frac{\text{precision} \times \text{recall}}{\text{precision} + \text{recall}} \]

On top of that, besides metrics, evaluating the calibration of the model is vital. Hence, two metrics were used in this paper, including Expected Calibration Error (ECE) and Maximum Calibration Error (MCE) with B=10 as the reference studies \cite{yuan2024does,naeini2015obtaining,niculescu2005predicting,nixon2019measuring}. They are formulated using the following equations:

\[
\text{ECE} = \sum_{i=1}^{B} P(i) \cdot |o_i - e_i|
\]

\[
\text{MCE} = \max_{i=1}^{B} (|o_i - e_i|)
\]

where:
\begin{itemize}
    \item \(o_i\) is the true fraction of positive instances in bin \(i\),
    \item \(e_i\) is the mean of the post-calibrated probabilities for the instances in bin \(i\),
    \item \(P(i)\) is the empirical probability (fraction) of all instances that fall into bin \(i\),
    \item \(B\) is the total number of bins.
\end{itemize}

The lower the values of ECE and MCE, the better the calibration of a model.

\subsection{Experimental Settings}
The experiments assessing ChatGPT's capability to diagnose AD in this research were conducted using OpenAI ChatGPT Version 4 Plus (GPT-4-turbo) \cite{openai2023a, achiam2023gpt}. Both zero-shot and multi-shot approaches were executed five times under three conditions: using MRI data alone, cognitive test scores alone, and a combination of MRI and cognitive test scores. In total, 30 runs were performed across both methods. To ensure independence between runs, all previous chat histories were cleared before initiating each new run. Regarding the examples for multi-shot prompting, 100 samples of each class were selected to put into the prompt. The thresholds are values of confidence scores that are equal or greater.

\begin{figure*}
    \centering
\includegraphics[width=0.9\linewidth]{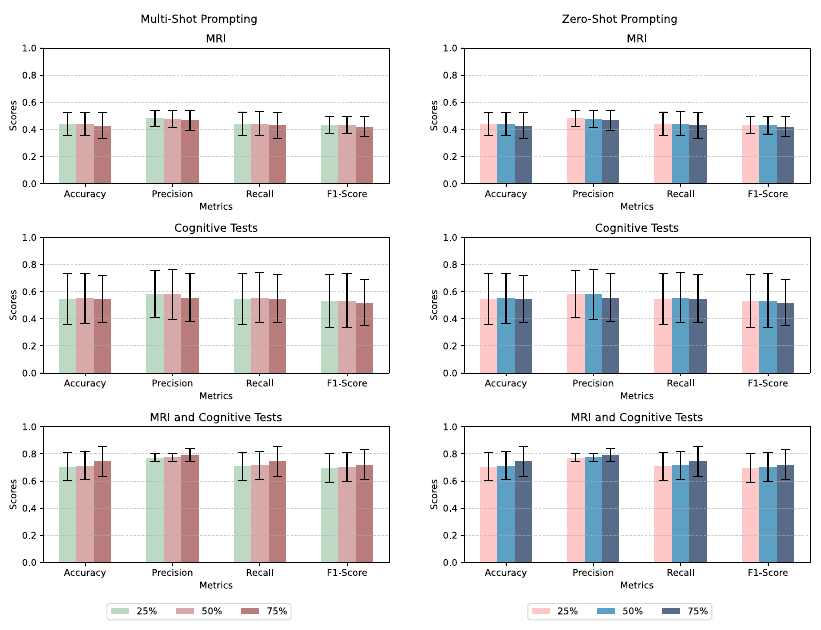}
    \caption{Visualisation of Performance Metrics of Zero-Shot and Multi-Shot Prompting with ChatGPT for Detecting AD.}
    \label{performace_metrics}
\end{figure*}

\begin{figure*}
    \centering
\includegraphics[width=0.6\linewidth]{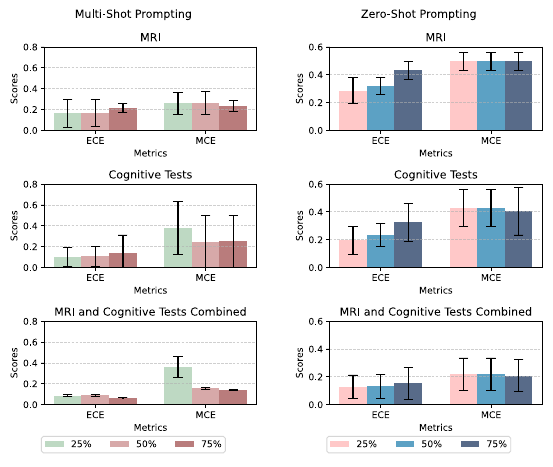}
    \caption{Visualisation of Calibration Metrics of Zero-Shot and Multi-Shot Prompting with ChatGPT for Detecting AD.}
    \label{ece_mce_metrics}
\end{figure*}

\section{Results}

\begin{figure*}
    \centering
\includegraphics[width=0.95\linewidth]{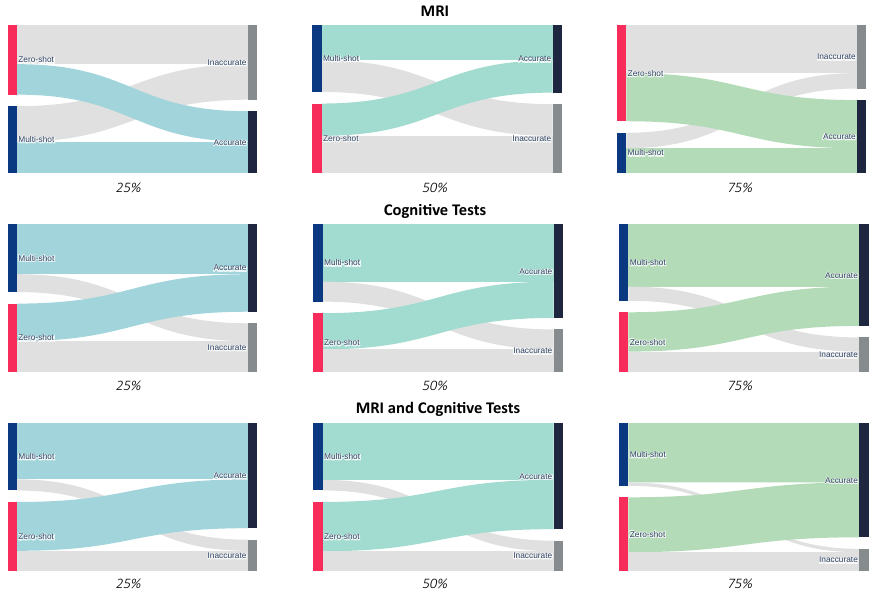}
    \caption{Accurate Samples with Different Thresholds from Zero-Shot and Multi-Shot Prompting for Detecting AD.}
    \label{sankey_results}
\end{figure*}

\begin{figure*}
    \centering
\includegraphics[width=0.95\linewidth]{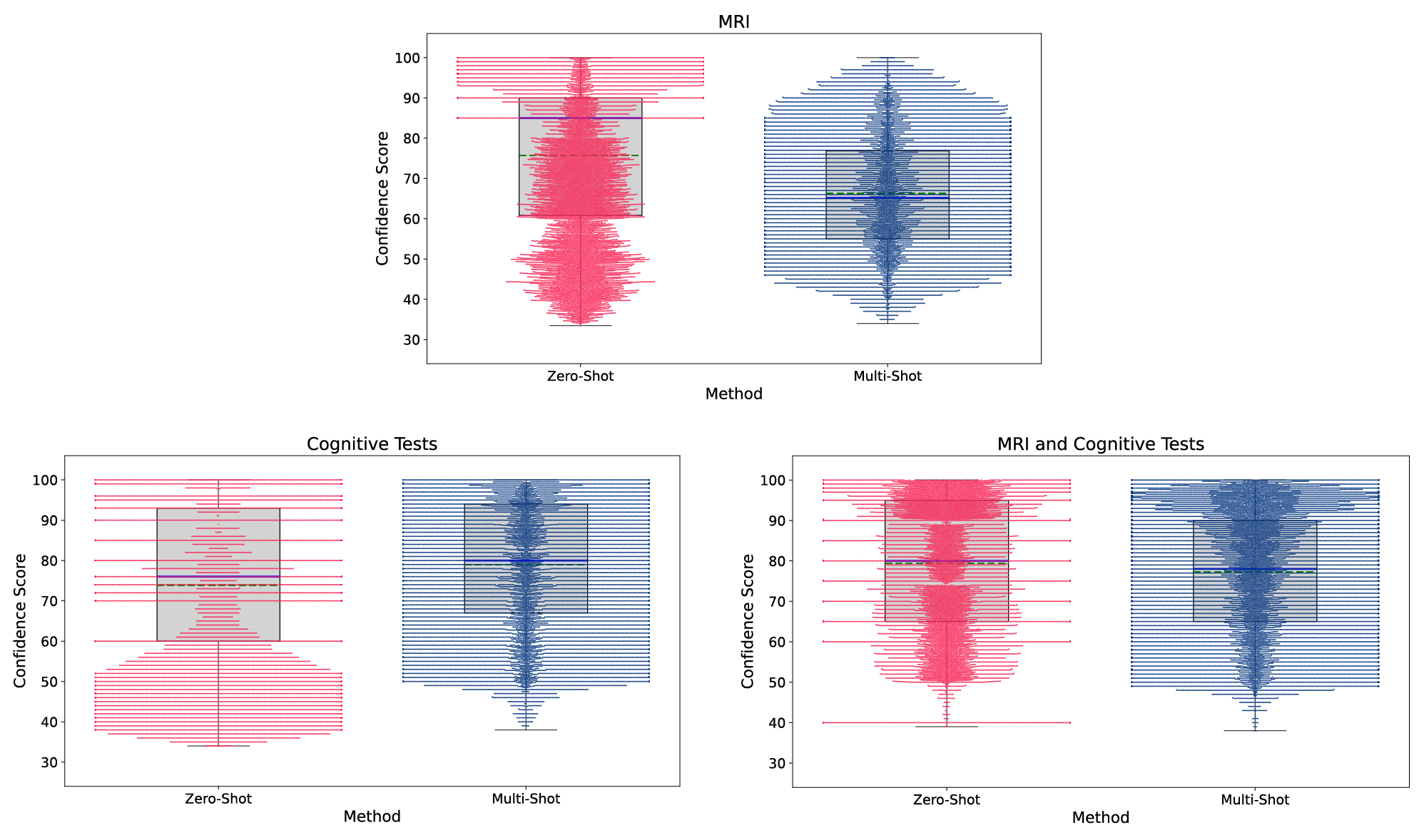}
    \caption{Accurate Samples with Confidence Scores (\%) Distribution from Zero-Shot and Multi-Shot Prompting for Detecting AD. \textit{The blue line represents the average, while the green dashed line is the median.}}
    \label{swarm_results}
\end{figure*}

\begin{table*}[ht]
\centering
\caption{Zero-Shot Prompting Results: Calibration Metrics.}
\begin{tabular}{|c|c|c|c|}
\hline
\textbf{Modality} & \textbf{Threshold} & \textbf{ECE $\downarrow$} & \textbf{MCE $\downarrow$} \\ \hline
\multirow{3}{*}{MRI} & 25\% & $0.284 \pm 0.093$ & $0.495 \pm 0.067$ \\ \cline{2-4}
                     & 50\% & $0.319 \pm 0.060$ & $0.495 \pm 0.067$ \\ \cline{2-4}
                     & 75\% & $0.433 \pm 0.065$ & $0.495 \pm 0.067$ \\ \hline
\multirow{3}{*}{Cognitive Tests} & 25\% & $0.194 \pm 0.100$ & $0.427 \pm 0.132$ \\ \cline{2-4}
                                 & 50\% & $0.233 \pm 0.085$ & $0.427 \pm 0.132$ \\ \cline{2-4}
                                 & 75\% & $0.325 \pm 0.135$ & $0.403 \pm 0.170$ \\ \hline
\multirow{3}{*}{MRI and Cognitive Tests} & 25\% & \textbf{0.129 $\pm$ 0.083} & \textit{0.220 $\pm$ 0.115} \\ \cline{2-4}
                                         & 50\% & \textit{0.131 $\pm$ 0.085} & \textit{0.220 $\pm$ 0.115} \\ \cline{2-4}
                                         & 75\% & $0.154 \pm 0.115$ & \textbf{0.207 $\pm$ 0.115} \\ \hline
\end{tabular}
\label{zero-shot-calibration-metrics}
\end{table*}

\begin{table*}[h!]
\centering
\caption{Zero-Shot Prompting Results: Performance Metrics.}
\begin{tabular}{|c|c|c|c|c|c|}
\hline
\textbf{Modality} & \textbf{Threshold} & \textbf{Accuracy $\uparrow$} & \textbf{Recall $\uparrow$} & \textbf{Precision $\uparrow$} & \textbf{F1-score $\uparrow$} \\ 
\hline
\multirow{3}{*}{MRI} 
    & 25\% & $0.440 \pm 0.085$ & $0.442 \pm 0.086$ & $0.481 \pm 0.059$ & $0.431 \pm 0.062$ \\ \cline{2-6}
    & 50\% & $0.439 \pm 0.087$ & $0.441 \pm 0.087$ & $0.477 \pm 0.064$ & $0.429 \pm 0.064$ \\ \cline{2-6}
    & 75\% & $0.427 \pm 0.094$ & $0.428 \pm 0.095$ & $0.466 \pm 0.074$ & $0.418 \pm 0.074$ \\ \hline
\multirow{3}{*}{Cognitive Tests} 
    & 25\% & $0.546 \pm 0.188$ & $0.548 \pm 0.189$ & $0.585 \pm 0.173$ & $0.533 \pm 0.195$ \\ \cline{2-6}
    & 50\% & $0.552 \pm 0.184$ & $0.554 \pm 0.185$ & $0.579 \pm 0.182$ & $0.534 \pm 0.197$ \\ \cline{2-6}
    & 75\% & $0.546 \pm 0.175$ & $0.547 \pm 0.177$ & $0.555 \pm 0.176$ & $0.519 \pm 0.167$ \\ \hline
\multirow{3}{*}{MRI and Cognitive Tests} 
    & 25\% & $0.706 \pm 0.104$ & $0.709 \pm 0.105$ & $0.771 \pm 0.031$ & $0.696 \pm 0.108$ \\ \cline{2-6}
    & 50\% & \textit{0.712 $\pm$ 0.103} & \textit{0.714 $\pm$ 0.104} & \textit{0.773 $\pm$ 0.030} & \textit{0.702 $\pm$ 0.107} \\ \cline{2-6}
    & 75\% & \textbf{0.744 $\pm$ 0.110} & \textbf{0.746 $\pm$ 0.111} & \textbf{0.791 $\pm$ 0.050} & \textbf{0.720 $\pm$ 0.112} \\ \hline
\end{tabular}
\label{zero-shor-peformance-metrics}
\end{table*}

\begin{table*}[ht]
\centering
\caption{Multi-Shot Prompting Results: Calibration Metrics.}
\begin{tabular}{|c|c|c|c|}
\hline
\textbf{Modality} & \textbf{Threshold} & \textbf{ECE $\downarrow$} & \textbf{MCE $\downarrow$} \\ \hline
\multirow{3}{*}{MRI} & 25\% & $0.162 \pm 0.136$ & $0.261 \pm 0.106$ \\ \cline{2-4}
                     & 50\% & $0.170 \pm 0.129$ & $0.262 \pm 0.108$ \\ \cline{2-4}
                     & 75\% & $0.214 \pm 0.040$ & $0.235 \pm 0.054$ \\ \hline
\multirow{3}{*}{Cognitive Tests} & 25\% & $0.104 \pm 0.092$ & $0.380 \pm 0.253$ \\ \cline{2-4}
                                 & 50\% & $0.105 \pm 0.093$ & $0.248 \pm 0.252$ \\ \cline{2-4}
                                 & 75\% & $0.143 \pm 0.168$ & $0.249 \pm 0.251$ \\ \hline
\multirow{3}{*}{MRI and Cognitive Tests} & 25\% & $0.087 \pm 0.010$ & $0.361 \pm 0.099$ \\ \cline{2-4}
                                         & 50\% & \textit{0.088 $\pm$ 0.010} & \textit{0.155 $\pm$ 0.010} \\ \cline{2-4}
                                         & 75\% & \textbf{0.066 $\pm$ 0.001} & \textbf{0.140 $\pm$ 0.005} \\ \hline
\end{tabular}
\label{multi-shot-prompting-calibration-results}
\end{table*}

\begin{table*}[h!]
\centering
\caption{Multi-Shot Prompting Results: Performance Metrics.}
\begin{tabular}{|c|c|c|c|c|c|}
\hline
\textbf{Modality} & \textbf{Threshold} & \textbf{Accuracy $\uparrow$} & \textbf{Recall $\uparrow$} & \textbf{Precision $\uparrow$} & \textbf{F1-score $\uparrow$} \\ 
\hline
\multirow{3}{*}{MRI} 
    & 25\% & $0.470 \pm 0.174$ & $0.470 \pm 0.174$ & $0.502 \pm 0.162$ & $0.441 \pm 0.204$ \\ \cline{2-6}
    & 50\% & $0.500 \pm 0.140$ & $0.500 \pm 0.140$ & $0.522 \pm 0.152$ & $0.455 \pm 0.184$ \\ \cline{2-6}
    & 75\% & $0.606 \pm 0.056$ & $0.606 \pm 0.056$ & $0.681 \pm 0.012$ & $0.510 \pm 0.090$ \\ \hline
\multirow{3}{*}{Cognitive Tests} 
    & 25\% & $0.736 \pm 0.137$ & $0.739 \pm 0.138$ & $0.749 \pm 0.121$ & $0.742 \pm 0.130$ \\ \cline{2-6}
    & 50\% & $0.742 \pm 0.138$ & $0.745 \pm 0.138$ & $0.754 \pm 0.123$ & $0.747 \pm 0.131$ \\ \cline{2-6}
    & 75\% & $0.809 \pm 0.199$ & $0.812 \pm 0.200$ & $0.815 \pm 0.190$ & $0.812 \pm 0.195$ \\ \hline
\multirow{3}{*}{MRI and Cognitive Tests} 
    & 25\% & $0.835 \pm 0.010$ & $0.838 \pm 0.010$ & $0.837 \pm 0.010$ & $0.837 \pm 0.011$ \\ \cline{2-6}
    & 50\% & \textit{0.843 $\pm$ 0.010} & \textit{0.846 $\pm$ 0.011} & \textit{0.845 $\pm$ 0.010} & \textit{0.844 $\pm$ 0.011} \\ \cline{2-6}
    & 75\% & \textbf{0.946 $\pm$ 0.001} & \textbf{0.950 $\pm$ 0.001} & \textbf{0.946 $\pm$ 0.001} & \textbf{0.948 $\pm$ 0.001} \\ \hline
\end{tabular}
\label{multi-shot-prompting-performance-results}
\end{table*}

\subsection{Zero-Shot Prompting}
To begin with, about the performance metrics of zero-shot prompting, as we can see in Table \ref{zero-shor-peformance-metrics} and Figures \ref{performace_metrics}, \ref{ece_mce_metrics} and \ref{swarm_results}. Firstly, combining MRI and cognitive test data yields superior performance across all metrics compared to using either modality alone. At a 75\% threshold, the combined modality achieves the highest accuracy ($0.744 \pm 0.110$), recall ($0.746 \pm 0.111$), precision ($0.791 \pm 0.050$), and F1-score ($0.720 \pm 0.112$), outperforming the individual modalities of MRI and cognitive tests. Notably, while cognitive tests alone provide better metrics than MRI alone, both unimodality show a decline in performance as the threshold increases, indicating that higher thresholds may limit the model's ability to perform effectively.

In terms of calibration, Table \ref{zero-shot-calibration-metrics} highlights that the combination of MRI and cognitive tests also delivers the most calibrated predictions, reflected by the lowest ECE and MCE. Specifically, at a 25\% threshold, the combined modality achieves an ECE of $0.129 \pm 0.083$ and an MCE of $0.220 \pm 0.115$, significantly better than using only MRI or cognitive tests' score. As thresholds increase, the calibration metrics for all modalities degrade slightly, with MRI exhibiting the highest ECE ($0.433 \pm 0.065$) at 75\%. These results underscore the value of integrating multiple data sources to improve both predictive performance and calibration.

\subsection{Multi-shot Prompting}
The performance and calibration metrics presented in Tables \ref{multi-shot-prompting-performance-results} and Figures \ref{performace_metrics}, \ref{ece_mce_metrics} and \ref{swarm_results} demonstrate the effectiveness of multi-shot prompting with ChatGPT for AD detection across different modalities and thresholds. From this table, it is noticeable that the integration of MRI and cognitive tests consistently outperforms individual modalities. At a 75\% threshold, the combined modality achieves the highest accuracy ($0.946 \pm 0.001$), recall ($0.950 \pm 0.001$), precision ($0.946 \pm 0.001$), and F1-score ($0.948 \pm 0.001$). This reflects the ability of multi-shot prompting to leverage complementary information from multiple data sources, leading to outstanding performance. The performance of using only MRI or cognitive tests' scores also demonstrates improvements, especially for cognitive tests, which reach an F1-score of $0.812 \pm 0.195$ at the 75\% threshold. However, MRI alone is left behind, particularly at lower thresholds, indicating its limited predictive power when not integrated with cognitive test data.

In terms of calibration, as we can see in Table \ref{multi-shot-prompting-calibration-results} the results further emphasize the advantages of multi-shot prompting. It indicates that the combination of MRI and cognitive tests yields the best-calibrated predictions, with an ECE of $0.066 \pm 0.001$ and an MCE of $0.140 \pm 0.005$ at the 75\% threshold. These values are significantly lower than those observed for using only MRI or cognitive tests, highlighting the effectiveness of the combined approach. Cognitive tests alone also exhibit strong calibration, particularly at lower thresholds, with an ECE of $0.104 \pm 0.092$ and an MCE of $0.380 \pm 0.253$ at the 25\% threshold. MRI, while showing improvement in calibration at higher thresholds, remains less calibrating compared to the others.

Overall, the results underscore the effectiveness of multi-shot prompting, particularly when utilizing multimodal data. The combination of MRI and cognitive tests not only improves performance metrics such as accuracy, precision, recall, and F1-score but also ensures better-calibrated predictions.

\subsection{Accurate Samples with Confidence Scores}
The analysis of accurately predicted samples, as illustrated in Figures \ref{sankey_results} and \ref{swarm_results}, reveals notable differences in the confidence scores across prompting methods and modalities. When using \textit{MRI data alone}, the \textit{average confidence score} for \textit{zero-shot prompting} is higher at \textit{85\%}, compared to \textit{multi-shot prompting}, which falls below \textit{70\%}. However, the \textit{median confidence score} for zero-shot is significantly lower---by more than \textit{10\%}---indicating greater variability and less consistency in its predictions. In contrast, for multi-shot prompting, the mean and median are closely aligned, suggesting a more stable and consistent distribution of confidence scores.

For \textit{cognitive tests only}, both methods exhibit relatively high confidence, but \textit{multi-shot prompting} outperforms zero-shot. The \textit{mean confidence score} for zero-shot is approximately \textit{76\%}, while multi-shot achieves a higher \textit{81\%}. This trend is similarly reflected in the \textit{median}, where multi-shot prompting exceeds zero-shot by around \textit{5\%}, indicating that multi-shot prompting achieves not only higher average confidence but also a more robust distribution.

Finally, when combining \textit{MRI and cognitive tests}, the confidence scores for \textit{zero-shot} and \textit{multi-shot prompting} are nearly equal. Both methods yield a \textit{mean confidence score} of approximately \textit{80\%} and a median of \textit{79\%} and \textit{78\%}, respectively. This suggests that integrating MRI and cognitive tests significantly improves prediction consistency, regardless of the prompting method. Overall, while \textit{zero-shot prompting} demonstrates higher confidence for MRI-only predictions, it comes with greater variability. In contrast, \textbf{multi-shot prompting} consistently delivers more stable confidence scores across all modalities, particularly excelling when cognitive tests or combined data are used. This highlights the advantage of multi-shot prompting in enhancing predictive confidence and minimizing uncertainty.

\section{Conclusion and Discussion}

First and foremost, based on the results developed with 9300 samples in this paper, we may conclude that ChatGPT can be a supportive tool to diagnose AD. However, there are some notices to leverage its capabilities. To begin with, this study addresses the research questions outlined in Section~\ref{RQ}, focusing on the emerging yet underexplored application of ChatGPT for AD detection. The findings demonstrate that ChatGPT can effectively diagnose AD using both zero-shot and multi-shot prompting approaches. Notably, combining \textit{MRI} and \textit{cognitive tests} as predictors outperform using either modality alone, highlighting the advantage of multimodal data integration.

When examining performance in detail, \textit{multi-shot prompting} significantly surpasses zero-shot prompting, achieving an accuracy of \textit{0.946} compared to \textit{0.744} for zero-shot. Both results were obtained with an optimal confidence threshold of \textit{75\%}. Furthermore, other performance metrics consistently tend to multi-shot prompting, underscoring its precision.

In addition, the calibration results strengthen the effectiveness of multi-shot prompting. Using combined MRI and cognitive tests, multi-shot prompting achieves ECE and MCE values of \textit{0.066} and \textit{0.005}, respectively, at a threshold of \textit{75\%}. While zero-shot prompting does not perform as well, it still demonstrates notable calibration improvements when combining MRI and cognitive tests. Specifically, zero-shot achieves its lowest ECE of \textit{0.129} at a \textit{25\%} threshold and an MCE of \textit{0.207} at a \textit{75\%} threshold.

Overall, these results highlight the clear advantage of multi-shot prompting for AD detection, both in predictive accuracy and calibration capability, particularly when leveraging the combined MRI and cognitive test data. This paper can open a new approach to AD detection, which is paramount for the QoL in societies \cite{lawton1994quality}. Especially it also provides opportunities for further research to leverage this technology for resource-limited regions in the world, to be a supportive tool easing the problem of shortage of AD specialists.

Regarding future developments of this research, several key objectives have been identified to enhance its scope and impact. Firstly, incorporating larger and more diverse datasets is essential to achieve more comprehensive and generalizable results, ensuring the robustness of the proposed approach. Secondly, conducting a fairness assessment \cite{zhang2023chatgpt,quttainah2024cost} is critical to evaluate potential biases in the model, particularly concerning its performance across different demographic groups or underprivileged populations. This will help address fairness concerns and ensure fair outcomes. Furthermore, ChatGPT should be compared with other techniques, such as Gemini or Llama 2 \cite{sandmann2024systematic,carla2024exploring}, to conduct a comparative evaluation and determine whether ChatGPT remains the most effective method for this application.

\bibliographystyle{ieeetr}
\bibliography{bib}

\end{document}